\documentclass[10pt, a4paper]{article}

\usepackage[final]{lrec2026} 

\usepackage{booktabs}
\usepackage{graphicx}
\usepackage{makecell}
\usepackage{arydshln}
\usepackage{multirow}
\usepackage{placeins}
\usepackage{array}

\title{QuantumBench: A Benchmark for Quantum Problem Solving}

\name{
    {\large\bf Shunya Minami\textsuperscript{1*}, Tatsuya Ishigaki\textsuperscript{1}, Ikko Hamamura\textsuperscript{2}}, \\
    {\large\bf Taku Mikuriya\textsuperscript{1,3}, Youmi Ma\textsuperscript{4}, Naoaki Okazaki\textsuperscript{4}}, \\
    {\large\bf Hiroya Takamura\textsuperscript{1}, Yohichi Suzuki\textsuperscript{1}, Tadashi Kadowaki\textsuperscript{1,5}}
}

\address{
    \textsuperscript{1}National Institute of Advanced Industrial Science and Technology (AIST) \\
    \textsuperscript{2}NVIDIA Corporation
    \textsuperscript{3}Yokohama National University
    \textsuperscript{4}Institute of Science Tokyo \\
    \textsuperscript{5}DENSO CORPORATION \\
    \textsuperscript{*}s-minami@aist.go.jp
}

\abstract{
Large language models are now integrated into many scientific workflows, accelerating data analysis, hypothesis generation, and design space exploration.
In parallel with this growth, there is a growing need to carefully evaluate whether models accurately capture domain-specific knowledge and notation, since general-purpose benchmarks rarely reflect these requirements.
This gap is especially clear in quantum science, which features non-intuitive phenomena and requires advanced mathematics.
In this study, we introduce QuantumBench, a benchmark for the quantum domain that systematically examine how well LLMs understand and can be applied to this non-intuitive field. 
Using publicly available materials, we compiled approximately 800 questions with their answers spanning nine areas related to quantum science and organized them into an eight-option multiple-choice dataset. 
With this benchmark, we evaluate several existing LLMs and analyze their performance in the quantum domain, including sensitivity to changes in question format. 
QuantumBench is the first LLM evaluation dataset built for the quantum domain, and it is intended to guide the effective use of LLMs in quantum research. 
 \\ \newline  
}

\begin{document}

\maketitleabstract

\section{Introduction}
\label{sec:intro}

Recent advances in large language models (LLMs) are reshaping scientific research activities. 
Beyond information gathering, paper summarization and manuscript drafting, LLMs are now used for hypothesis generation and experimental design. 
Furthermore, recent studies are beginning to demonstrate the utility of research-agent systems for end-to-end research workflows, including AI Scientist~\cite{lu2024ai, yamada2025ai} and Agent Laboratory~\cite{schmidgall2025agent}.
These developments indicate a plausible path toward greater automation of the research pipeline and raise the prospect of AI-enabled scientific discoveries.

For example, in quantum science, LLMs are applied to tasks ranging from code generation to experimental design. 
The Qiskit Code Assistant~\cite{dupuis2024qiskit} can generate code for quantum programming with the Qiskit library~\cite{qiskit2024}, and is integrated into the IBM Quantum platform. 
Similarly, El Agente Q~\cite{zou2025agente} autonomously conducts quantum chemistry calculations by orchestrating more than twenty domain-specific agents under LLM supervision. 
With natural-language instructions alone, users can prepare input files and execute computations, enabling non-specialists to perform advanced quantum chemistry simulations.
In addition, the agent-based AI framework called k-agents automates quantum-computing experiments by planning, executing, and analyzing them using LLMs~\cite{cao2025automating}.

However, the performance of LLMs on scientific tasks has not yet been sufficiently validated.
LLMs can generate plans with serious errors and inconsistencies, such as fabricated references or datasets, as well as proposals for physically infeasible experimental conditions~\cite{eger2025transforming}. 
In quantum many-body physics, for example, LLMs perform well on prescribed computational procedures but show limited ability to incorporate physical context when planning experiments~\cite{pan2025quantum}. 
These observations imply that LLMs do not yet possess robust, domain-specific scientific reasoning, underscoring the need to assess whether they genuinely acquire domain knowledge.

In this work, we focus on quantum science, a domain marked by non-intuitive phenomena and advanced mathematical structures, where existing general-purpose benchmarks are likely insufficient to evaluate LLM performance adequately.
To address this challenge, we construct QuantumBench, a benchmark dataset that systematically assesses how well LLMs understand and can be applied to quantum science. 
Using publicly available resources such as MIT OpenCourseWare~\cite{MITOCW}, TU Delft OpenCourseWare~\cite{TUDOCW}, and LibreTexts~\cite{LibreTexts}, we have selected fifteen courses and textbooks related to quantum science and collected the corresponding lecture notes, exercises, and assignments.
From these materials, question–answer pairs have been created, and encoded as eight-option multiple-choice questions with plausible but incorrect options alongside the correct answer.
QuantumBench comprises approximately 800 undergraduate-level question–answer pairs across nine quantum-related subfields.

QuantumBench enables comprehensive, comparative evaluation of existing LLMs in the quantum domain. 
We quantify LLM performance on quantum tasks, including sensitivity to variation in question format and wording. 
Beyond quantum science, the benchmark highlights the need for further improvements in the use of LLMs across scientific fields and helps accelerate the development of AI for scientific research.

\section{Related Work}
\label{sec:related}

Various general-purpose benchmarks exist.
For example, The Massive Multitask Language Understanding (MMLU) benchmark~\cite{hendrycks2020measuring} is a widely used benchmark of the general capabilities of LLMs across academic and professional domains. 
It comprises multiple-choice questions from 57 subjects, including elementary mathematics and computer science, and assesses broad knowledge relevant to research. 
MMLU-Pro~\cite{wang2024mmlu} extends MMLU by increasing the number of answer choices to ten to raise reasoning difficulty. 
For advanced evaluations, GPQA~\cite{rein2024gpqa} is proposed as a representative benchmark targeting PhD-level subjects with expert-authored, high-quality problems that are not answerable through simple search or shallow reasoning, making it a standard for assessing doctoral-level competence. 
More recently, Humanity’s Last Exam~\cite{phan2025humanity} spans dozens of fields across mathematics, the humanities, and the natural sciences, and includes 3,000 ultra-challenging problems that demand deep reasoning and advanced expertise beyond data lookup.

Numerous domain-specific benchmark datasets have also been proposed. 
For mathematics, MATH~\cite{hendrycks2021measuring} contains about 12{,}500 problems drawn from competitive contests in the United States, spanning algebra, number theory, geometry, probability, and combinatorics. 
Unlike general benchmarks such as MMLU and GPQA, which primarily use multiple-choice formats, MATH employs short-answer, free-response questions without predefined options, thereby evaluating precise mathematical reasoning and solution derivation. 
In the biomedical domain, PubMedQA~\cite{jin2019pubmedqa} is a representative benchmark, containing approximately 1,000 questions in the form of “Yes/No/Maybe,” created based on academic questions from papers published in PubMed.
PubMedQA is widely used to evaluate large language models’ ability to understand and interpret medical contexts, thereby facilitating advancements in biomedical research and drug discovery.
Collectively, such domain-specific benchmarks probe specialized abilities and sensitivity to the representations and conventions of each field that general-purpose evaluations do not capture, enabling more valid assessments of model performance in real-world settings.

In contrast, benchmark datasets specialized for the quantum domain remain limited. Within GPQA, only 64 items cover quantum mechanics. 
More recently, CritPt~\cite{zhu2025probing} was introduced to test whether models can perform advanced scientific reasoning required in frontier physics; it comprises 71 challenges with 190 intermediate reasoning checkpoints, including 15 questions in quantum information science and technology. 
While these resources are valuable for assessing capabilities across science and physics, they are insufficient for comprehensive benchmarking of model ability in quantum science.

For quantum-specific evaluation, major efforts focus on coding skill for quantum programming, with benchmarks such as QiskitHumanEval~\cite{vishwakarma2024qiskit} and QCoder Benchmark~\cite{mikuriya2025qcoder}. 
As a corpus of question–answer pairs, QuantumLLMInstruct (QLMMI)~\cite{kashani2024quantumllminstruct} includes 500{,}000 items. 
However, since QLMMI was generated entirely by LLMs, and designed for fine-tuning rather than rigorous evaluation, it does not serve as an adequate benchmark for assessing model performance.
Human-authored datasets such as MMLU and GPQA remain important for advancing applications. 
In this study, we construct a benchmark from publicly available resources by collecting undergraduate-level questions authored by domain experts, thereby providing an evaluation framework for knowledge understanding and reasoning in the quantum domain.

\section{QuantumBench}
In this section, we outline the construction procedure of QuantumBench and present an overview of the dataset.

\begin{table*}[tb] 
\resizebox{\textwidth}{!}{%
\begin{tabular}{lllr}
\toprule
Platform & Lecture Titles & Domain & Number of Problems \\
\midrule
MIT OCW & Computational Quantum Mechanics of Molecular and Extended Systems & Quantum Chemistry & 2\\
MIT OCW & Introduction to Applied Nuclear Physics & Nuclear Physics & 38 \\
MIT OCW & Introductory Quantum Mechanics I & Quantum Mechanics & 5 \\
MIT OCW & Optics (2014) & Optics & 49 \\
MIT OCW & Optics (2009) & Optics & 122 \\
MIT OCW & Quantum Computation & Quantum Computation & 43 \\
MIT OCW & Quantum Optical Communication & Photonics & 118 \\
MIT OCW & Quantum Physics I & Quantum Mechanics & 130 \\
MIT OCW & Relativistic Quantum Field Theory I & Quantum Field Theory & 81 \\
MIT OCW & Relativistic Quantum Field Theory II & Quantum Field Theory & 21 \\
MIT OCW & Relativistic Quantum Field Theory III & Quantum Field Theory & 8 \\
MIT OCW & String Theory and Holographic Duality & String Theory & 23 \\
MIT OCW & String Theory & String Theory & 17 \\
TUD OCW & Quantum Information Processing & Quantum Computation & 19 \\
LibreTexts & Physical Chemistry I, Quantum Mechanics & Quantum Chemistry & 93 \\
\bottomrule
\end{tabular}%
}
\caption{The titles of the lectures used as data sources, their corresponding academic fields, and the number of questions collected} 
\label{tab:qa-source}
\end{table*}

\begin{table}[tb]
\centering
\scalebox{0.65}{%
\begin{tabular}{lrrrr}
\toprule
& \makecell{Algebraic\\Calculation} 
& \makecell{Numerical\\Calculation} 
& \makecell{Conceptual\\Understanding} 
& Total \\
\midrule
Quantum Mechanics & 177 & 21 & 14 & 212 \\
Quantum Computation & 54 & 1 & 5 & 60 \\
Quantum Chemistry & 16 & 64 & 6 & 86 \\
Quantum Field Theory & 104 & 1 & 2 & 107 \\
Photonics & 54 & 1 & 2 & 57 \\
Mathematics & 37 & 0 & 0 & 37 \\
Optics & 101 & 41 & 15 & 157 \\
Nuclear Physics & 1 & 15 & 2 & 18 \\
String Theory & 31 & 0 & 2 & 33 \\
\midrule
Total & 575 & 144 & 50 & 769 \\
\bottomrule
\end{tabular}}
\caption{Number of questions per domains}
\label{tab:qa-number}
\end{table}

\subsection{Dataset Construction}

We selected 15 quantum science–related online courses and teaching materials from MIT OpenCourseWare, TU Delft OpenCourseWare, and LibreTexts. 
These sources include assignments and examinations together with instructor-provided solutions. 
From these materials, we chose 769 questions for which the question–answer mapping was unambiguous and the solution was uniquely determined. 
Table~\ref{tab:qa-source} summarizes the course titles, subject areas, and the number of collected items.

Next, we preprocessed the collected question statements. 
To ensure that each question can be understood and solved in isolation, we manually augmented the texts with missing information, including implicit notation and unstated definitions, with assistance from an LLM (gpt-oss-120b~\cite{agarwal2025gpt}).
We also used the LLM to flag ambiguities or omissions and to check internal consistency and completeness. 
For each problem, we created seven plausible but incorrect options in addition to the correct answer.
These options were human-curated and cover not only simple mistakes, such as sign errors, but also responses reflecting misinterpretation of the statement or mistakes arising at intermediate calculations.

Finally, some of the authors annotated each problem with one of three question types: algebraic calculation (symbolic manipulation and formula derivation), numerical calculation (computations with explicit numerical values), and conceptual understanding (knowledge- and principle-based questions without explicit calculation). 
We also assigned a field label to each problem. 
Because some courses include questions from related areas (for example, a quantum mechanics course may include a few quantum chemistry questions), we relabeled fields based on actual content to improve the quality of categorization. 
The category comprises nine fields in total: eight lecture-oriented fields plus an additional Mathematics category.

We also categorized all 769 problems by difficulty and expertise level. 
Three researchers in quantum science performed the level annotation, rating difficulty level on a five-point scale from Level 1 (trivial) to Level 5 (challenging) and expertise level on a four-point scale from Level 1 (high school) to Level 4 (PhD). 
It is acceptable that the annotators do not assign levels to some items because those topics were outside their areas of expertise.
For each problem, the final label was the median of the three ratings. 
The criteria appear in Table~\ref{tab:qa-difficulty} and Table~\ref{tab:qa-expertise}.

\subsection{Summary of the Dataset}

Table~\ref{tab:qa-number} shows the number of items in the final dataset by fields. 
Of the 769 problems, Algebraic Calculation is the largest category, with 575 items (approximately 75\% of the dataset), indicating that the collection is dominated by questions focused on symbolic and analytical derivation. 
By subfield, quantum mechanics accounts for the largest share, followed by optics.
Several representative examples are presented in Table~\ref{tab:display_questions}.

\begin{figure}[t]
    \centering
    \includegraphics[clip, width=0.9\columnwidth]{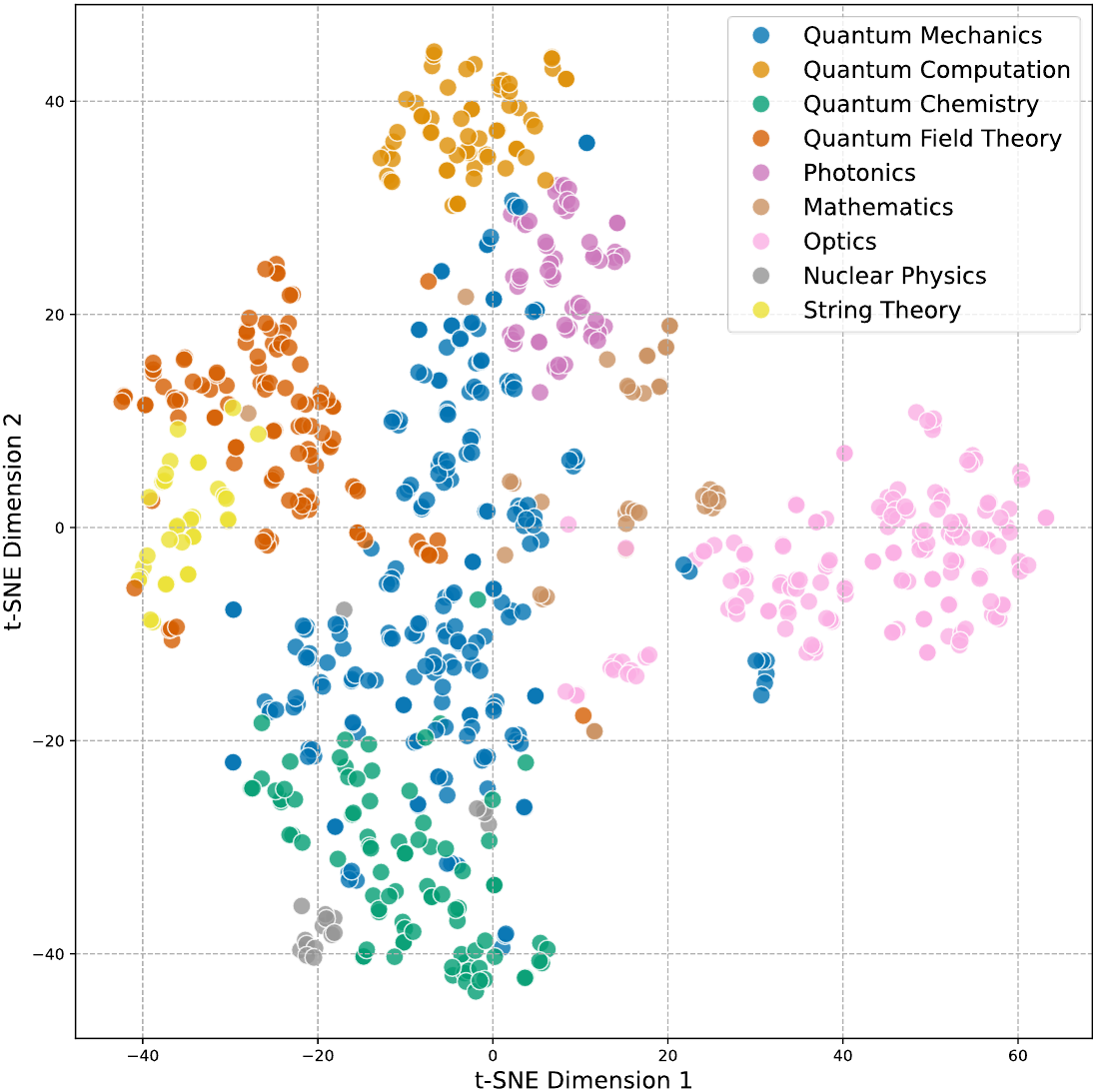}
    \caption{
    Visualization of QuantumBench. 
    The questions were embedded into 3072-dimensional vectors using text-embedding-3-large, and mapped onto a two-dimensional space using t-SNE.
    }
    \label{fig:tsne}
\end{figure}

Figure~\ref{fig:tsne} shows a t-SNE~\cite{maaten2008visualizing} projection of the question statements embedded with OpenAI’s text-embedding-3-large model.
The resulting clusters align with fields, indicating clear semantic separation in the embedding space.
The cluster structure is roughly organized into four groups: (i) Quantum Computation, (ii) Optics, (iii) Quantum Mechanics and related areas including Quantum Chemistry and Nuclear Physics, and (iv) Quantum Field Theory and String Theory, which require advanced algebraic calculations.

\begin{table}[tb] 
\resizebox{0.5\textwidth}{!}{%
\begin{tabular}{ll}
\toprule
 & Criteria  \\
\midrule
Level 1 & A problem whose correct answer can be obtained immediately \\
Level 2 & A problem with an obvious solution \\
 & that can be solved with simple calculations \\
Level 3 & A problem whose solution comes to mind quickly \\
 & but requires somewhat tedious steps \\
Level 4 & A problem that requires some thought to discover the solution, \\
                         & or whose solution is obvious but involves considerably tedious steps \\
Level 5 & A problem whose solution cannot be easily identified \\
\bottomrule
\end{tabular}
}
\caption{Difficulty levels and their criteria} 
\label{tab:qa-difficulty}
\end{table}
\begin{table}[tb] 
\resizebox{0.5\textwidth}{!}{%
\begin{tabular}{ll}
\toprule
 & Criteria  \\
\midrule
Level 1 & An elementary problem, non-specialists can understand the question. \\
Level 2 & People who studied physics can understand the the question. \\
Level 3 & Understanding requires having read technical texts in the field. \\
Level 4 & Only experts who conduct research in that field can understand the question. \\
\bottomrule
\end{tabular}
}
\caption{Expertise levels and their criteria} 
\label{tab:qa-expertise}
\end{table}

\begin{figure}[t]
    \centering
    \includegraphics[clip, width=0.70\columnwidth]{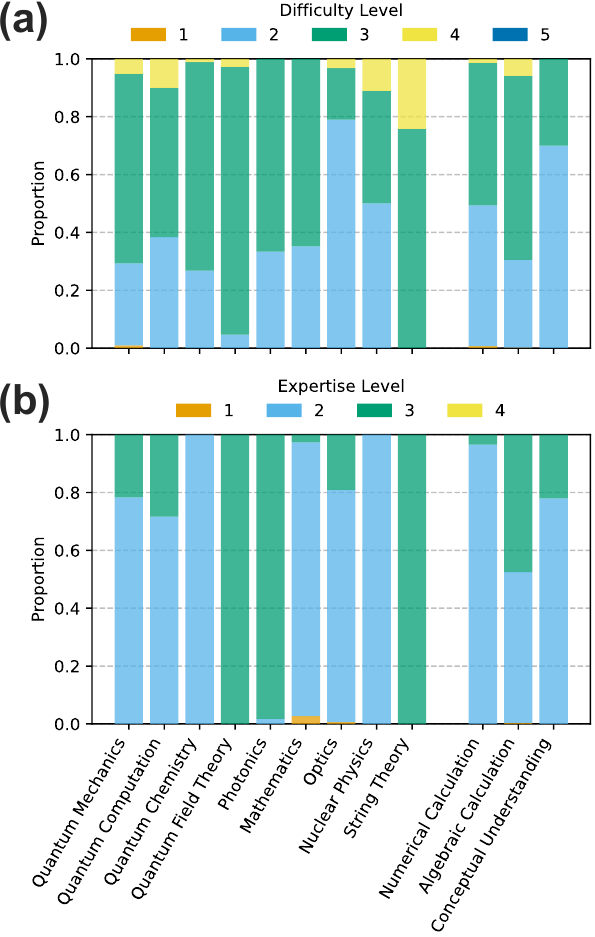}
    \caption{
    Distribution of difficulty and expertise levels by category.
    (a) Difficulty level, ranging from 1 (trivial) to 5 (challenging).
    (b) Expertise level, ranging from 1 (high school) to 4 (PhD).
    The graph shows the distribution of problems that were evaluated by at least one annotator.
    }
    \label{fig:diff-exp}
\end{figure}

Figure~\ref{fig:diff-exp} illustrates the distribution of difficulty and expertise level across the categories (domain and type). 
The overall average difficulty level is 2.68, and the average expertise level is 2.37. 
These statistics suggest that QuantumBench primarily comprises standard-level questions typical of undergraduate education.
Notably, no problems classified as difficulty level 5 (challenging) or requiring an expertise level 4 (PhD) are included.
This indicates that QuantumBench serves as a benchmark to evaluate whether a model can consistently demonstrate the fundamental knowledge and reasoning abilities required for conducting research in quantum science.

Because the levels of difficulty and expertise are not uniform among domains and problem types, cross-category comparisons within a single model are not meaningful. 
Instead, QuantumBench should be used to evaluate model performance independently within each category, enabling fair comparisons of distinct models.

%
%
%
%

\begin{table*}
    \centering
    \small
    {\fontsize{8}{9}\selectfont
        \begin{tabularx}{\textwidth}{X}
            \toprule
            \multicolumn{1}{c}{\textbf{Quantum Computation}} \\
            \midrule
            Consider the state $\frac{1}{\sqrt{2}}(|000\rangle + |111\rangle)$. Suppose all three qubits of this state are measured in the $|+\rangle = \{\frac{1}{\sqrt{2}}(|0\rangle + |1\rangle), |-\rangle=\frac{1}{\sqrt{2}}(|0\rangle - |1\rangle)\}$ basis. What are the possible joint outcomes of these measurements? \\
            \vspace{1pt}
            \begin{tabular}{@{}p{0.48\textwidth} p{0.48\textwidth}@{}}
                (A) $\frac{1}{\sqrt{2}}(|-+-\rangle + |+-+\rangle)$ & (E) $\frac{1}{\sqrt{2}}(|+--\rangle + |-++\rangle)$ \\
                (B) $\frac{1}{2}(|+++\rangle + |++-\rangle + |-+-\rangle + |+--\rangle)$ & (F) $\frac{1}{\sqrt{2}}(|--+\rangle + |++-\rangle)$ \\
                (C) $\frac{1}{2}(|+++\rangle + |++-\rangle + |---\rangle + |-++\rangle)$ & (G) $\frac{1}{2}(|+++\rangle + |--+\rangle + |-+-\rangle + |+--\rangle)$ \\
                (D) $\frac{1}{\sqrt{2}}(|+++\rangle + |---\rangle)$ & (H) $\frac{1}{2}(|+++\rangle + |--+\rangle + |---\rangle + |-++\rangle)$ \\
            \end{tabular} \\
            \midrule
            \multicolumn{1}{c}{\textbf{Quantum Field Theory}} \\
            \midrule
            In this problem we consider the chiral representation, and write a Dirac spinor $\psi$ in terms of two chiral spinors $\psi_L$ and $\psi_R$ as
            $
            \psi = [ \psi_L , \psi_R ]^\top.
            $
            The Lagrangian density for the Dirac theory contains a mass term of the form
            $$
            \mathcal{L} = \dots + im\bar{\psi}\psi = \dots + im(\psi_L^{\dagger}\psi_R + \psi_R^{\dagger}\psi_L).
            $$
            \vspace{-1pt}
            \textbf{Statements}
            \begin{itemize}
            \item[1.] The above mass term is Lorentz invariant.
            \item[2.] $m\psi_L^{\dagger}\psi_L$ is Lorents invariant.
            \item[3.] $m\psi_R^{\dagger}\psi_R$ is Lorents invariant.
            \end{itemize} \\
            \vspace{0.1pt}
            \begin{tabular}{p{0.23\textwidth}@{\hspace{2pt}}p{0.23\textwidth}@{\hspace{2pt}}p{0.23\textwidth}@{\hspace{2pt}}p{0.23\textwidth}}
            (A) 1. True, 2. True, 3. True &
            (B) 1. True, 2. True, 3. False &
            (C) 1. True, 2. False, 3. True &
            (D) 1. True, 2. False, 3. False \\
            (E) 1. False, 2. True, 3. True &
            (F) 1. False, 2. False, 3. False &
            (G) 1. False, 2. True, 3. False &
            (H) 1. False, 2. False, 3. True \\
            \end{tabular} \\
            \midrule
            \multicolumn{1}{c}{\textbf{Optics}} \\
            \midrule
            The Lloyd mirror is a wavefront-splitting interferometer. It consists of a flat glass mirror that reflects a portion of wavefront that comes from a narrow slit. Another portion of the wavefront proceeds directly to the screen. The interference of the two wavefronts form a set of bright and dark fringes that can be measured on the screen.

            In our problem, let's assume the mirror is placed at the plane $x = 0$ and illuminated by a spherical wave originating from the slit at location $(x_0, -z_0)$ (where $x_0, z_0 > 0$). Using the paraxial approximation for a 1D spherical wave $(y=0)$,
            $$
            E(x, z) = E_0 \frac{\exp[i(k(z + z_0)]}{i(z + z_0)} \exp\left[ik\frac{(x - x_0)^2}{2(z + z_0)}\right]
            $$
            The source illuminating the slit has a wavelength of 500nm in air. If the slit is positioned at $x_0$ =1mm above the flat mirror, and the screen is placed 1 meter away from the slit, please estimate the spacing of the fringes on the screen. \\
            \vspace{1pt}
            \begin{tabular}{p{0.23\textwidth}@{\hspace{2pt}}p{0.23\textwidth}@{\hspace{2pt}}p{0.23\textwidth}@{\hspace{2pt}}p{0.23\textwidth}}
                (A) 25 mm & 
                (B) 500 mm &
                (C) 0.75 mm &
                (D) 100 mm \\
                (E) 0.25 mm &
                (F) 250 mm &
                (G) 0.50 mm &
                (H) 1.00 mm \\
            \end{tabular} \\
            \bottomrule
        \end{tabularx}
    }
    \caption{Examples of questions included in QuantumBench. From three fields, examples are given for algebraic calculation, conceptual understanding, and numerical calculation.}
    \label{tab:display_questions}
\end{table*}

\begin{figure*}[t]
    \centering
    \includegraphics[clip, width=2\columnwidth]{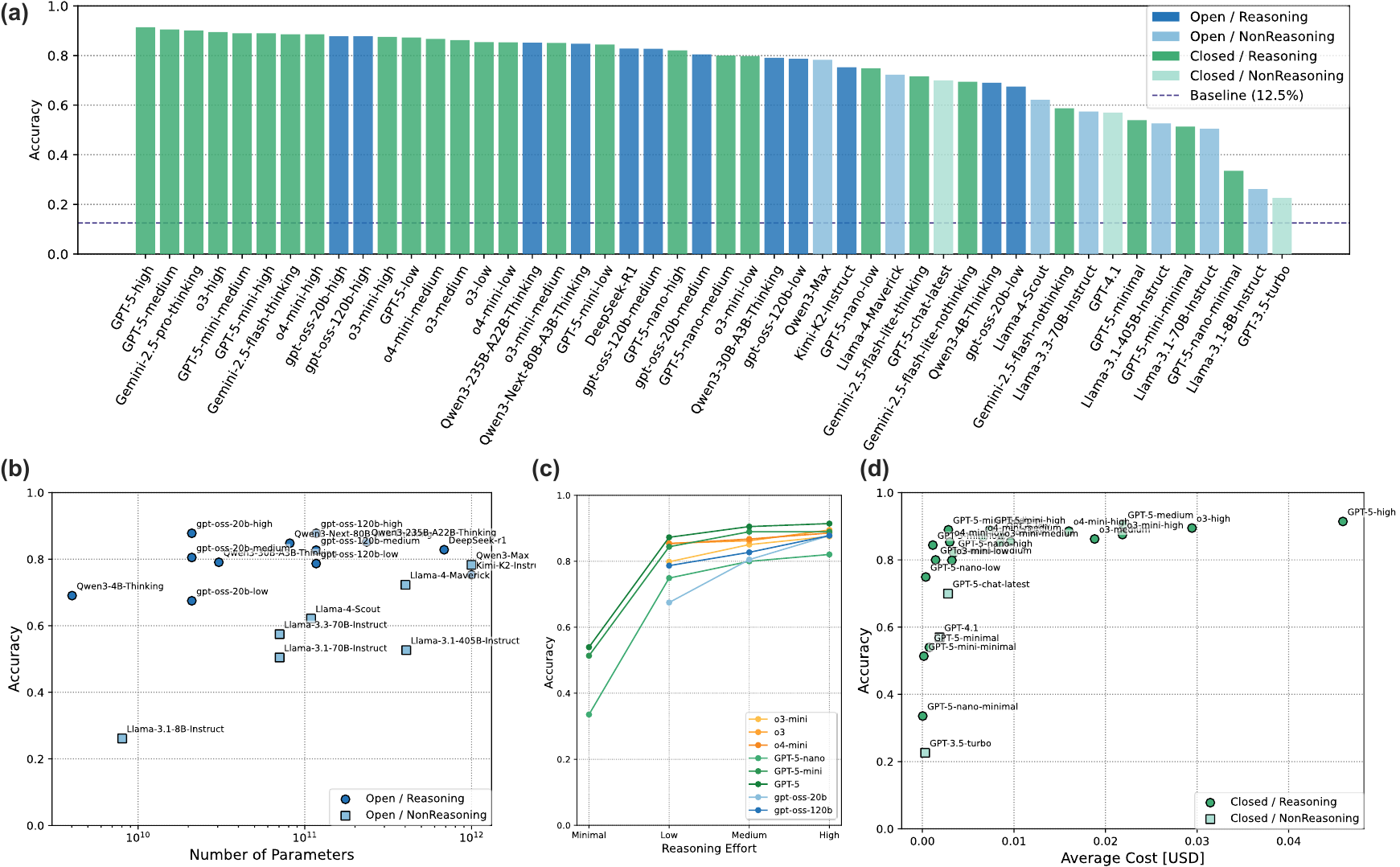}
    \caption{
    Benchmark results.  
    (a) Accuracy on all 769 problems. Blue indicates open-weight models, and green indicates closed models available only via API. Dark colors denote reasoning models, while light colors denote non-reasoning models. Tags appended to the names of reasoning models indicate the strength of reasoning.  
    (b) Relationship between the number of model parameters and accuracy. Only open-weight models with publicly available parameter counts are shown. Blue circles represent reasoning models, and light-blue squares represent non-reasoning models.  
    (c) Transition of accuracy when varying the reasoning strength for reasoning models. The “minimal” setting is available only for some models.  
    (d) Relationship between average API usage cost per problem and accuracy. Only closed models, which require API usage, are shown. Green circles represent reasoning models, and light-green circles represent non-reasoning models.  
    }
    \label{fig:accuracy}
\end{figure*}

\section{Experiments}

We evaluated 
11 OpenAI models~\cite{brown2020language, achiam2023gpt, agarwal2025gpt}, 
6 Meta models~\cite{touvron2023llama, dubey2024llama}, 
5 Alibaba models~\cite{yang2025qwen3}, 
3 Google models~\cite{team2023gemini}, 
1 DeepSeek model~\cite{guo2025deepseek}, 
and 1 Moonshot AI model~\cite{team2025kimi}.
For OpenAI and Google models without public weights, inference was conducted via API; all other models were run on NVIDIA H100 GPUs using ABCI-Q. 
For reasoning models, we varied reasoning strength and evaluated multiple configurations. 
During inference, web search and external tool use were disabled to evaluate the isolate model-internal problem-solving ability.
Our benchmark data and code is available at GitHub\footnote{\url{https://github.com/shunyaist/QuantumBench}}.

\subsection{Zero-Shot}

For each problem, the model received only the question statement and answer choices, and we evaluated whether it could select the correct answer~\cite{bowman2022measuring}.
Figure~\ref{fig:accuracy}(a) reports accuracy for all 48 model configurations on the full QuantumBench dataset. 
The current frontier model, GPT-5, achieved the highest performance, while open-weight models such as gpt-oss-120b and Qwen3-235B reached comparable accuracy. 
Most high-performing models were reasoning models. 
Although non-reasoning models, including Qwen3-Max and Llama4-Maverik, performed relatively well, their accuracy did not exceed 80\%. 
These results indicate that many tasks in the quantum domain require multi-step reasoning, and that explicit mechanisms for sequential reasoning are a key determinant of performance.

Model accuracy generally increases with the number of parameters~\cite{kaplan2020scaling}. 
Figure~\ref{fig:accuracy}(b) shows the relationship between model size and accuracy, and the similar trend holds in QuantumBench. 
Non-reasoning models exhibit a clear performance gain as scale grows. 
By contrast, reasoning models achieve accuracy comparable to large non-reasoning models even with fewer parameters. 
Figure~\ref{fig:accuracy}(c) further confirms the role of reasoning by plotting performance as a function of reasoning strength from minimal to high.
Deeper reasoning consistently improves accuracy, with a marked drop at the minimal setting and substantial gains even at the low setting.

This perspective is especially relevant to cost effectiveness. 
Figure~\ref{fig:accuracy}(d) shows the relationship between API inference cost and accuracy. 
Notably, performance gains diminish sharply as cost increases. 
In other words, even small models, under medium reasoning and low-cost settings, can approach the accuracy of frontier models. 
These findings indicate that, for AI tools supporting scientific research, adopting small- to medium-scale models with moderate reasoning capabilities as the default configuration can achieve an effective balance between performance and computational cost.

\begin{figure*}[t]
    \centering
    \includegraphics[clip, width=1.99\columnwidth]{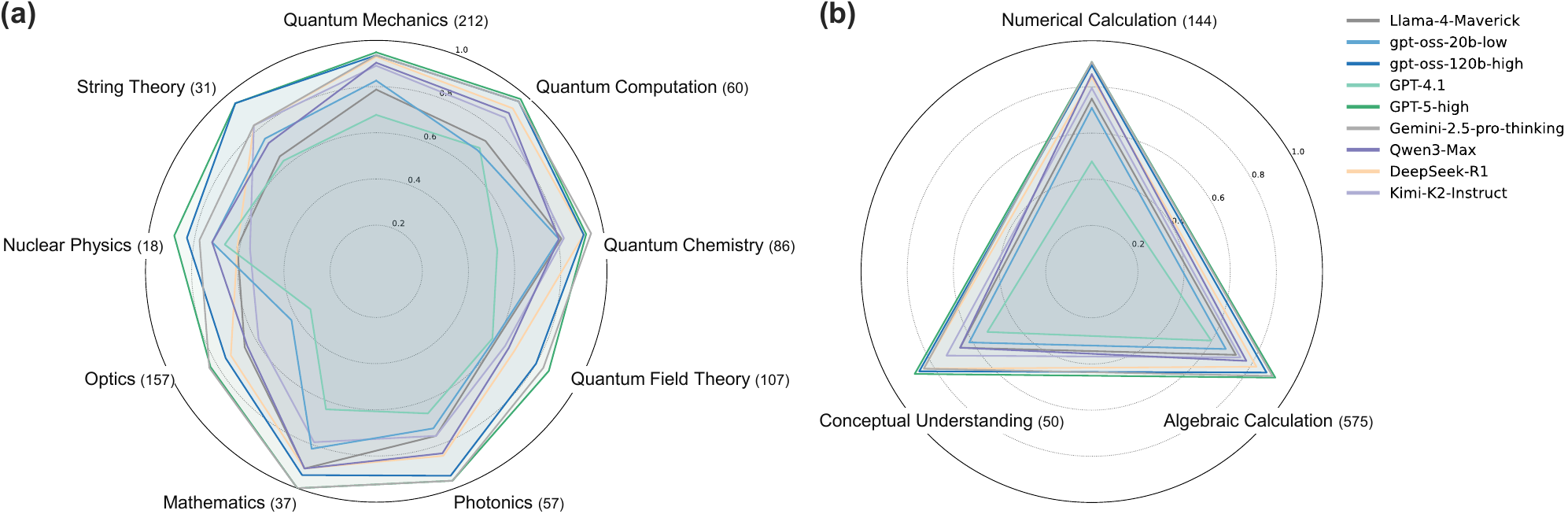}
    \caption{
    Category-wise accuracy of the LLMs.  
    (a) Accuracy by question domain. The numbers attached to the axis labels indicate the number of questions in each domain.  
    (b) Accuracy by question type.  
    }
    \label{fig:radar}
\end{figure*}

Figure~\ref{fig:radar} illustrates accuracy by question category for some frontier and mid-tier models from each provider. 
In Optics, prior or non-reasoning models achieved only limited accuracy, whereas large reasoning models approached 80\%. 
Many Optics questions rely on diagrammatic or spatial information such as lens placement, relative spacing, and propagation direction, which is difficult to process linguistically and is considered one factor underlying this performance gap.
Similarly, fields such as String Theory, Quantum Field Theory, and Nuclear Physics demand relatively advanced domain knowledge and reasoning, resulting in a larger variance in accuracy across models.

Figure~\ref{fig:radar}(b) shows accuracy by question type. 
There were no notable behavioral differences among the models across problem types. 
GPT-5-high, which achieved the highest overall accuracy, consistently outperformed the other models across all three question types.

\begin{figure}[t]
    \centering
    \includegraphics[clip, width=0.99\columnwidth]{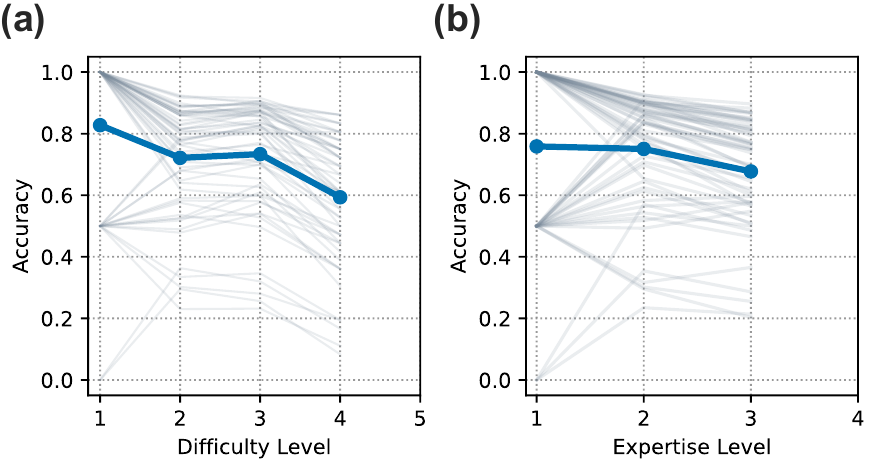}
    \caption{
    Accuracy of LLMs across (a) difficulty and (b) expertise levels. Each gray line shows the accuracy of an individual model, and the blue line shows the average accuracy across all models.
    }
    \label{fig:acc-vs-level}
\end{figure}

Figure~\ref{fig:acc-vs-level} shows the relationship between model accuracy and both difficulty and expertise levels. 
In both categories, accuracy declined progressively as the level increased. 
However, the decrease was relatively moderate, and no statistically significant differences in performance were observed between difficulty levels 2 (easy) and 3 (standard) or between expertise levels 1 (high school) and 2 (undergraduate).
These findings suggest that errors on problems of medium-level difficulty or expertise are driven by factors other than difficulty or expertise alone. 
Section~\ref{sec:error} provides a detailed analysis of these error sources.

\subsection{Zero-Shot CoT}

\begin{figure}[t]
    \centering
    \includegraphics[clip, width=0.7\columnwidth]{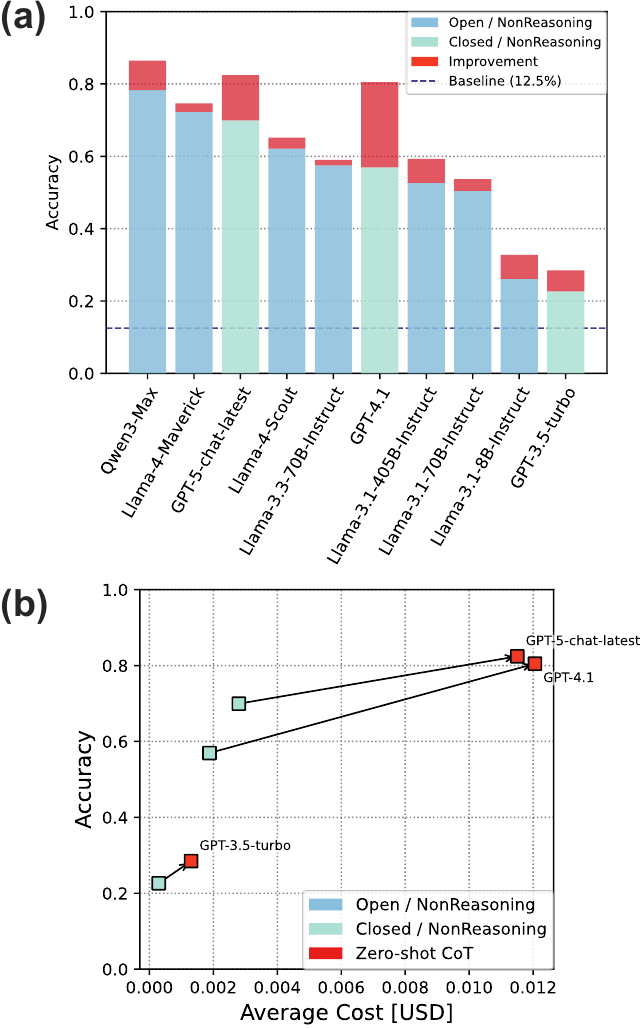}
    \caption{
    Performance changes with zero-shot CoT.  
    (a) Accuracy of non-reasoning models in the zero-shot setting, and the improvement with zero-shot CoT. Blue indicates open-weight models, and green indicates closed models. The red area represents the improvement margin due to zero-shot CoT.  
    (b) Transition of accuracy and inference cost when zero-shot CoT is applied.  
    }
    \label{fig:cot}
\end{figure}

Zero-shot CoT~\cite{kojima2022large} is a prompting method that directs the model to perform chain-of-thought (CoT) reasoning without examples, thereby inducing step-by-step inference. 
We applied zero-shot CoT to ten non-reasoning models and examined the resulting performance improvements.

Figure~\ref{fig:cot}(a) shows the accuracy gains obtained with zero-shot CoT. 
Although accuracy improved for many models, the magnitude of improvement was generally limited. 
Some of high-performing models, such as Qwen3-Max, GPT-5-chat-latest, and GPT-4.1 showed the most pronounced gains, whereas many prior models with low base performance exhibited only minor improvements. 
This indicates that models lacking sufficient baseline ability cannot achieve substantial improvement through simple prompting-based reasoning techniques.
The notable improvement observed in GPT-4.1 is likely due to its ability to process very long contexts (exceeding one million tokens), allowing for extended reasoning steps.

As models are prompted to perform deeper reasoning, the number of generated tokens and associated inference cost increase. 
Figure~\ref{fig:cot}(b) illustrates this rise in inference cost when zero-shot CoT is applied. GPT-4.1, which achieved the largest accuracy gain, also incurred a significant cost increase due to the generation of long-form reasoning traces. 
In contrast, GPT-5-chat-latest also showed an increase in cost, while the performance improvement was limited. 
This observation implies that questions that remain unsolved by high-performing models under zero-shot settings cannot be resolved simply by extending elementary reasoning procedures.
In other words, current models still struggle to construct effective reasoning chains for complex scientific questions. 
Because many quantum-domain tasks require long step-by-step theoretical analyses, models must be trained to develop robust long-form reasoning steps.

\subsection{Error Analysis}
\label{sec:error}
\begin{figure}[t]
    \centering
    \includegraphics[clip, width=0.93\columnwidth]{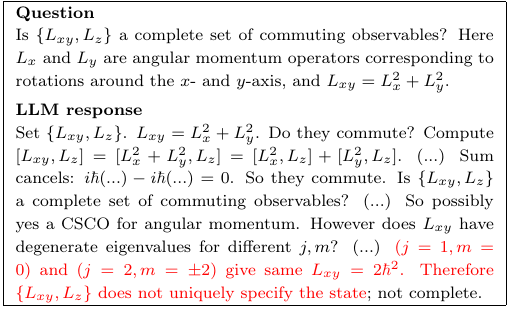}
    \caption{
    An example of incorrect responses by LLMs.
    We extract a snippet of the question and the reasoning process.
    The red text indicates the errors in the reasoning.
    The LLM claims incompleteness by presenting an example where $L_{xy}$ takes the same value, but since $L_z$ differs, the example is not a valid counterexample.
    }
    \label{fig:error-example}
\end{figure}

Systematic analysis of common or characteristic error patterns in LLMs supports high-accuracy deployment in practical applications.
Figure~\ref{fig:error-example} presents a representative case in which the experts rated the question as level 2 (easy), yet the average accuracy across the evaluated LLMs was only about 29.2\%. 
The figure also displays the reasoning process produced by gpt-oss-120b-high. 
The task asks whether two operators constitute a complete set of commuting observables (CSCO). 
A correct solution requires verifying a sequence of necessary conditions and then concluding that the operators form a CSCO. 
The model skipped part of this verification and therefore produced the incorrect conclusion that they do not.

Failures to perform the necessary reasoning steps in scientific contexts constitute one of the primary sources of error on QuantumBench and have also been documented in the prior work~\cite{pan2025quantum}. 
Other frequent causes include excessive reliance on common sense at the expense of the stated definitions and assumptions, errors in handling indices and signs, and failure to follow instructions, which often leads the model to report a different target quantity. 
Systematically mitigating these tendencies is essential for building more accurate and trustworthy AI systems.

%

\subsection{Bias Analysis for LLM-as-a-Judge}
We examined whether an LLM can appropriately judge the level of difficulty and expertise. 
We used GPT-5-chat-latest to assign difficulty and expertise levels to each QuantumBench question and compared these labels with human annotations. 
The decision criteria embedded in the prompts matched those in Tables~\ref{tab:qa-difficulty} and \ref{tab:qa-expertise}. 

\begin{figure}[t]
    \centering
    \includegraphics[clip, width=1\columnwidth]{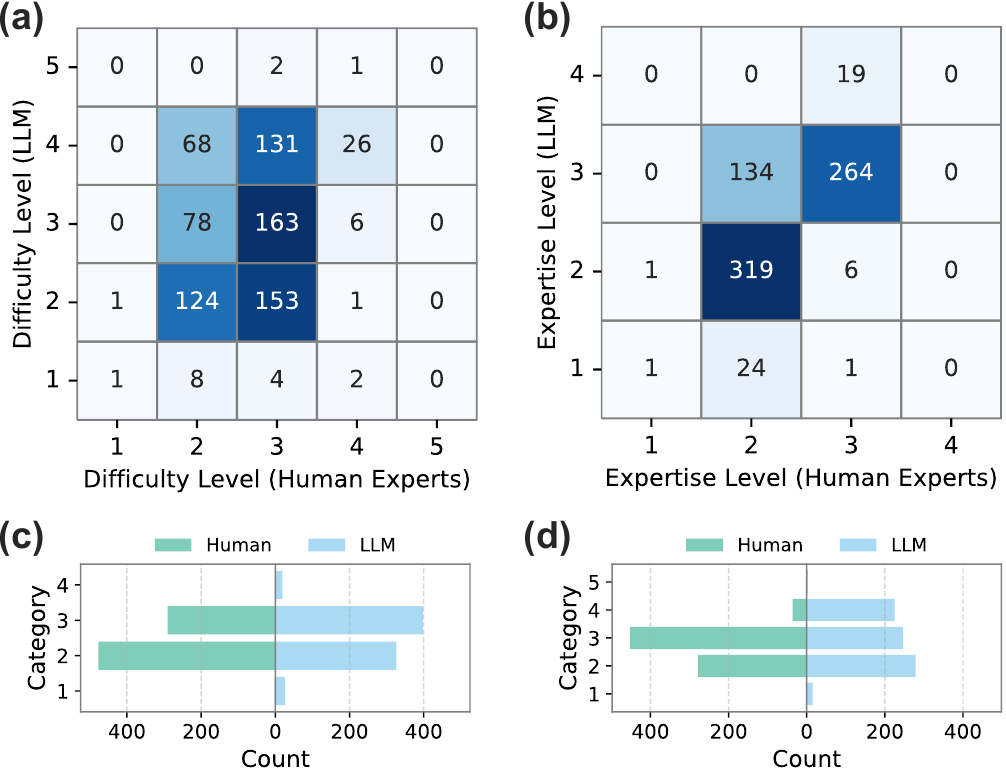}
    \caption{
    Differences in difficulty and expertise ratings between humans and the LLM.
    (a, c) Heatmap comparing human and LLM ratings.
    (b, d) Distribution of human and LLM ratings.
    }
    \label{fig:human-vs-llm}
\end{figure}

Figure~\ref{fig:human-vs-llm} summarizes the correspondence between LLM and human labels.
In the difficulty annotation task, the LLM rated the questions as more difficult than human experts: while humans placed most questions at Level 2 (easy) or Level 3 (standard), the LLM’s labels were roughly uniform across Levels 2–4. 
A similar trend held in the expertise annotation task: the LLM tended to judge the questions as requiring higher levels of expertise than human experts.

Several factors may account for these discrepancies. 
First, while the LLM appears to recognize that most problems do not fall into the extreme difficulty levels (Levels 1 and 5), it may have failed to reliably distinguish among the intermediate categories (Levels 2–4). 
Consequently, the model likely assigned labels within that range uniformly at random. 
In addition, multiple biases have been noted when using LLMs as judges. 
For example, a verbosity bias has been observed, in which longer responses tend to be rated more favorably~\cite{zheng2023judging, ye2024justice}. 
In the evaluation of mathematical content, judgments often reflect stylistic features rather than substantive correctness~\cite{stephan2024calculation}. 
Given that many QuantumBench items contain detailed problem descriptions and numerous equations, these characteristics may have amplified such biases, leading the LLM to overestimate the difficulty and degree of expertise required even for fundamentally elementary problems.

\section{Conclusion}

In this study, we introduced QuantumBench, a benchmark for evaluating LLMs in the quantum domain, constructed from publicly available lecture materials. 
The dataset comprises 769 undergraduate-level multiple-choice problems across nine areas spanning core and related quantum fields. 
Using this benchmark, we evaluated existing models, including current frontier models, and analyzed performance by model type, reasoning strength, and computational cost. 
We found that small to medium models can approach the accuracy of frontier models when prompted with moderate reasoning. 
We also observed that performance improvements diminish with increasing reasoning strength and clarified that practical performance can be attained even under low-cost settings.

QuantumBench is the first benchmark specialized for LLM evaluation in the quantum domain and supports the advancement of LLM use in quantum science. 
However, because the dataset is drawn from undergraduate and graduate course materials, it does not fully capture the difficulty, quality, or formats characteristic of cutting-edge research and development. 
Many real-world research tasks are not multiple choice but instead require goal specification, procedural decomposition, verification, and interpretation. 
To develop more sophisticated AI scientists, future benchmarks should include more practical tasks such as open-ended descriptive questions, structured task decomposition, and systematic experiment planning.

\section{Limitations}
Because the dataset was constructed from course materials that are publicly available on the web, the underlying sources and related resources may have been included in the training corpora of LLMs. 
In most of these sources, however, problems and solutions are distributed in separate files, which makes it difficult for an LLM to memorize exact problem-answer pairs. 
In addition, we have edited the problem statements and newly created all choices. 
These processes mitigate training‑data contamination and keep the dataset suitable for evaluation. 

We further evaluated model performance on a Japanese version of the benchmark and found that results were very similar to those on the English original, indicating that models are not merely recalling memorized question–answer pairs.

Because enlisting domain experts across all subfields in the dataset was infeasible, we did not measure human accuracy on the benchmark. 
Instead, the physics experts assigned categorical ratings for difficulty and expertise level. 
From the standpoint of evaluating LLM capabilities, establishing a human baseline remains desirable and is deferred to future work.

\section{Ethics Statements}

\subsection{License}
All source lecture materials were distributed under the CC BY-NC-SA license, and the dataset created in this study is distributed under the same license. 
The title of each source lecture, the lecturer’s name, and the lecture URL are explicitly included as metadata within the dataset.

\subsection{Human Annotations}

The annotation was conducted by three of the authors. 
Prior to the task, we held a meeting to share the details of the annotation process and provided a comprehensive annotation guideline to standardize the evaluation criteria and working procedures. 
The annotators understood and agreed that their work would be released as part of the dataset. 
During the annotation process, we communicated with the annotators as needed to discuss how to handle irregular cases.

\section{Acknowledgements}
This work was performed for Council for Science, Technology and Innovation (CSTI), Cross-ministerial Strategic Innovation Promotion Program (SIP), “Promoting the application of advanced quantum technology platforms to social issues” (Funding agency : QST).
We used ABCI-Q provided by AIST with support from “ABCI-Q Special Project Support Usage".

\section{Bibliographical References}\label{sec:reference}

\bibliographystyle{lrec2026-natbib}
\bibliography{lrec2026}

\begin{thebibliography}{38}
\expandafter\ifx\csname natexlab\endcsname\relax\def\natexlab#1{#1}\fi

\bibitem[{Achiam et~al.(2023)Achiam, Adler, Agarwal, Ahmad, Akkaya, Aleman, Almeida, Altenschmidt, Altman, Anadkat et~al.}]{achiam2023gpt}
Josh Achiam, Steven Adler, Sandhini Agarwal, Lama Ahmad, Ilge Akkaya, Florencia~Leoni Aleman, Diogo Almeida, Janko Altenschmidt, Sam Altman, Shyamal Anadkat, et~al. 2023.
\newblock {GPT}-4 technical report.
\newblock \emph{arXiv preprint arXiv:2303.08774}.

\bibitem[{Agarwal et~al.(2025)Agarwal, Ahmad, Ai, Altman, Applebaum, Arbus, Arora, Bai, Baker, Bao et~al.}]{agarwal2025gpt}
Sandhini Agarwal, Lama Ahmad, Jason Ai, Sam Altman, Andy Applebaum, Edwin Arbus, Rahul~K Arora, Yu~Bai, Bowen Baker, Haiming Bao, et~al. 2025.
\newblock gpt-oss-120b \& gpt-oss-20b model card.
\newblock \emph{arXiv preprint arXiv:2508.10925}.

\bibitem[{Bowman et~al.(2022)Bowman, Hyun, Perez, Chen, Pettit, Heiner, Luko{\v{s}}i{\=u}t{\.e}, Askell, Jones, Chen et~al.}]{bowman2022measuring}
Samuel~R Bowman, Jeeyoon Hyun, Ethan Perez, Edwin Chen, Craig Pettit, Scott Heiner, Kamil{\.e} Luko{\v{s}}i{\=u}t{\.e}, Amanda Askell, Andy Jones, Anna Chen, et~al. 2022.
\newblock Measuring progress on scalable oversight for large language models.
\newblock \emph{arXiv preprint arXiv:2211.03540}.

\bibitem[{Brown et~al.(2020)Brown, Mann, Ryder, Subbiah, Kaplan, Dhariwal, Neelakantan, Shyam, Sastry, Askell et~al.}]{brown2020language}
Tom Brown, Benjamin Mann, Nick Ryder, Melanie Subbiah, Jared~D Kaplan, Prafulla Dhariwal, Arvind Neelakantan, Pranav Shyam, Girish Sastry, Amanda Askell, et~al. 2020.
\newblock Language models are few-shot learners.
\newblock \emph{Advances in neural information processing systems}, 33:1877--1901.

\bibitem[{Cao et~al.(2025)Cao, Zhang, Alghadeer, Fasciati, Piscitelli, Bakr, Leek, and Aspuru-Guzik}]{cao2025automating}
Shuxiang Cao, Zijian Zhang, Mohammed Alghadeer, Simone~D Fasciati, Michele Piscitelli, Mustafa Bakr, Peter Leek, and Al{\'a}n Aspuru-Guzik. 2025.
\newblock Automating quantum computing laboratory experiments with an agent-based {AI} framework.
\newblock \emph{Patterns}.

\bibitem[{{Delft University of Technology}()}]{TUDOCW}
{Delft University of Technology}.
\newblock {TU} {Delft} {OpenCourseWare}.
\newblock \url{https://ocw.tudelft.nl/}.

\bibitem[{Dubey et~al.(2024)Dubey, Jauhri, Pandey, Kadian, Al-Dahle, Letman, Mathur, Schelten, Yang, Fan et~al.}]{dubey2024llama}
Abhimanyu Dubey, Abhinav Jauhri, Abhinav Pandey, Abhishek Kadian, Ahmad Al-Dahle, Aiesha Letman, Akhil Mathur, Alan Schelten, Amy Yang, Angela Fan, et~al. 2024.
\newblock The {L}lama 3 herd of models.
\newblock \emph{arXiv e-prints}, pages arXiv--2407.

\bibitem[{Dupuis et~al.(2024)Dupuis, Buratti, Vishwakarma, Forrat, Kremer, Faro, Puri, and Cruz-Benito}]{dupuis2024qiskit}
Nicolas Dupuis, Luca Buratti, Sanjay Vishwakarma, Aitana~Viudes Forrat, David Kremer, Ismael Faro, Ruchir Puri, and Juan Cruz-Benito. 2024.
\newblock Qiskit code assistant: Training {LLM}s for generating quantum computing code.
\newblock In \emph{2024 IEEE LLM Aided Design Workshop (LAD)}, pages 1--4. IEEE.

\bibitem[{Eger et~al.(2025)Eger, Cao, D'Souza, Geiger, Greisinger, Gross, Hou, Krenn, Lauscher, Li et~al.}]{eger2025transforming}
Steffen Eger, Yong Cao, Jennifer D'Souza, Andreas Geiger, Christian Greisinger, Stephanie Gross, Yufang Hou, Brigitte Krenn, Anne Lauscher, Yizhi Li, et~al. 2025.
\newblock Transforming science with large language models: A survey on {AI}-assisted scientific discovery, experimentation, content generation, and evaluation.
\newblock \emph{arXiv preprint arXiv:2502.05151}.

\bibitem[{{Gemini Team}(2023)}]{team2023gemini}
{Gemini Team}. 2023.
\newblock Gemini: a family of highly capable multimodal models.
\newblock \emph{arXiv preprint arXiv:2312.11805}.

\bibitem[{Guo et~al.(2025)Guo, Yang, Zhang, Song, Zhang, Xu, Zhu, Ma, Wang, Bi et~al.}]{guo2025deepseek}
Daya Guo, Dejian Yang, Haowei Zhang, Junxiao Song, Ruoyu Zhang, Runxin Xu, Qihao Zhu, Shirong Ma, Peiyi Wang, Xiao Bi, et~al. 2025.
\newblock Deep{S}eek-{R}1: Incentivizing reasoning capability in {LLM}s via reinforcement learning.
\newblock \emph{arXiv preprint arXiv:2501.12948}.

\bibitem[{Hendrycks et~al.(2020)Hendrycks, Burns, Basart, Zou, Mazeika, Song, and Steinhardt}]{hendrycks2020measuring}
Dan Hendrycks, Collin Burns, Steven Basart, Andy Zou, Mantas Mazeika, Dawn Song, and Jacob Steinhardt. 2020.
\newblock Measuring massive multitask language understanding.
\newblock \emph{arXiv preprint arXiv:2009.03300}.

\bibitem[{Hendrycks et~al.(2021)Hendrycks, Burns, Kadavath, Arora, Basart, Tang, Song, and Steinhardt}]{hendrycks2021measuring}
Dan Hendrycks, Collin Burns, Saurav Kadavath, Akul Arora, Steven Basart, Eric Tang, Dawn Song, and Jacob Steinhardt. 2021.
\newblock Measuring mathematical problem solving with the math dataset.
\newblock \emph{arXiv preprint arXiv:2103.03874}.

\bibitem[{Javadi-Abhari et~al.(2024)Javadi-Abhari, Treinish, Krsulich, Wood, Lishman, Gacon, Martiel, Nation, Bishop, Cross, Johnson, and Gambetta}]{qiskit2024}
Ali Javadi-Abhari, Matthew Treinish, Kevin Krsulich, Christopher~J. Wood, Jake Lishman, Julien Gacon, Simon Martiel, Paul~D. Nation, Lev~S. Bishop, Andrew~W. Cross, Blake~R. Johnson, and Jay~M. Gambetta. 2024.
\newblock \href {https://doi.org/10.48550/arXiv.2405.08810} {Quantum computing with {Q}iskit}.
\newblock \emph{arXiv preprint arXiv:2405.08810}.

\bibitem[{Jin et~al.(2019)Jin, Dhingra, Liu, Cohen, and Lu}]{jin2019pubmedqa}
Qiao Jin, Bhuwan Dhingra, Zhengping Liu, William~W Cohen, and Xinghua Lu. 2019.
\newblock {PubMedQA}: A dataset for biomedical research question answering.
\newblock \emph{arXiv preprint arXiv:1909.06146}.

\bibitem[{Kaplan et~al.(2020)Kaplan, McCandlish, Henighan, Brown, Chess, Child, Gray, Radford, Wu, and Amodei}]{kaplan2020scaling}
Jared Kaplan, Sam McCandlish, Tom Henighan, Tom~B Brown, Benjamin Chess, Rewon Child, Scott Gray, Alec Radford, Jeffrey Wu, and Dario Amodei. 2020.
\newblock Scaling laws for neural language models.
\newblock \emph{arXiv preprint arXiv:2001.08361}.

\bibitem[{Kashani(2024)}]{kashani2024quantumllminstruct}
Shlomo Kashani. 2024.
\newblock {QuantumLLMInstruct}: A 500k {LLM} instruction-tuning dataset with problem-solution pairs for quantum computing.
\newblock \emph{arXiv preprint arXiv:2412.20956}.

\bibitem[{{Kimi Team}(2025)}]{team2025kimi}
{Kimi Team}. 2025.
\newblock Kimi {K}2: Open agentic intelligence.
\newblock \emph{arXiv preprint arXiv:2507.20534}.

\bibitem[{Kojima et~al.(2022)Kojima, Gu, Reid, Matsuo, and Iwasawa}]{kojima2022large}
Takeshi Kojima, Shixiang~Shane Gu, Machel Reid, Yutaka Matsuo, and Yusuke Iwasawa. 2022.
\newblock Large language models are zero-shot reasoners.
\newblock \emph{Advances in neural information processing systems}, 35:22199--22213.

\bibitem[{{L}ibre{T}exts()}]{LibreTexts}
{L}ibre{T}exts.
\newblock {L}ibre{T}exts.
\newblock \url{https://libretexts.org/}.

\bibitem[{Lu et~al.(2024)Lu, Lu, Lange, Foerster, Clune, and Ha}]{lu2024ai}
Chris Lu, Cong Lu, Robert~Tjarko Lange, Jakob Foerster, Jeff Clune, and David Ha. 2024.
\newblock The ai scientist: Towards fully automated open-ended scientific discovery.
\newblock \emph{arXiv preprint arXiv:2408.06292}.

\bibitem[{Maaten and Hinton(2008)}]{maaten2008visualizing}
Laurens van~der Maaten and Geoffrey Hinton. 2008.
\newblock Visualizing data using t-{SNE}.
\newblock \emph{Journal of machine learning research}, 9(Nov):2579--2605.

\bibitem[{{Massachusetts Institute of Technology}()}]{MITOCW}
{Massachusetts Institute of Technology}.
\newblock {MIT} {OpenCourseWare}.
\newblock \url{https://ocw.mit.edu/}.

\bibitem[{Mikuriya et~al.(2025)Mikuriya, Ishigaki, Minami, Kadowaki, Suzuki, Naito, Takada, Kato, Baseda, Yamadaand, and Takamura}]{mikuriya2025qcoder}
Taku Mikuriya, Tatsuya Ishigaki, Shunya Minami, Tadashi Kadowaki, Yohichi Suzuki, Shun Naito, Shunya Takada, Takumi Kato, Tamotsu Baseda, Reo Yamadaand, and Hiroya Takamura. 2025.
\newblock {QCoder} {B}enchmark: Bridging language generation and quantum hardware through simulator-based feedback.
\newblock In \emph{Proceedings of the 18th International Natural Language Generation Conference}.

\bibitem[{Pan et~al.(2025)Pan, Mudur, Taranto, Tikhanovskaya, Venugopalan, Bahri, Brenner, and Kim}]{pan2025quantum}
Haining Pan, Nayantara Mudur, William Taranto, Maria Tikhanovskaya, Subhashini Venugopalan, Yasaman Bahri, Michael~P Brenner, and Eun-Ah Kim. 2025.
\newblock Quantum many-body physics calculations with large language models.
\newblock \emph{Communications Physics}, 8(1):49.

\bibitem[{Phan et~al.(2025)Phan, Gatti, Han, Li, Hu, Zhang, Zhang, Shaaban, Ling, Shi et~al.}]{phan2025humanity}
Long Phan, Alice Gatti, Ziwen Han, Nathaniel Li, Josephina Hu, Hugh Zhang, Chen Bo~Calvin Zhang, Mohamed Shaaban, John Ling, Sean Shi, et~al. 2025.
\newblock Humanity's last exam.
\newblock \emph{arXiv preprint arXiv:2501.14249}.

\bibitem[{Rein et~al.(2024)Rein, Hou, Stickland, Petty, Pang, Dirani, Michael, and Bowman}]{rein2024gpqa}
David Rein, Betty~Li Hou, Asa~Cooper Stickland, Jackson Petty, Richard~Yuanzhe Pang, Julien Dirani, Julian Michael, and Samuel~R Bowman. 2024.
\newblock {GPQA}: A graduate-level google-proof {Q}\&{A} benchmark.
\newblock In \emph{First Conference on Language Modeling}.

\bibitem[{Schmidgall et~al.(2025)Schmidgall, Su, Wang, Sun, Wu, Yu, Liu, Moor, Liu, and Barsoum}]{schmidgall2025agent}
Samuel Schmidgall, Yusheng Su, Ze~Wang, Ximeng Sun, Jialian Wu, Xiaodong Yu, Jiang Liu, Michael Moor, Zicheng Liu, and Emad Barsoum. 2025.
\newblock Agent laboratory: Using {LLM} agents as research assistants.
\newblock \emph{arXiv preprint arXiv:2501.04227}.

\bibitem[{Stephan et~al.(2024)Stephan, Zhu, A{\ss}enmacher, Shen, and Roth}]{stephan2024calculation}
Andreas Stephan, Dawei Zhu, Matthias A{\ss}enmacher, Xiaoyu Shen, and Benjamin Roth. 2024.
\newblock From calculation to adjudication: Examining {LLM} judges on mathematical reasoning tasks.
\newblock \emph{arXiv preprint arXiv:2409.04168}.

\bibitem[{Touvron et~al.(2023)Touvron, Lavril, Izacard, Martinet, Lachaux, Lacroix, Rozi{\`e}re, Goyal, Hambro, Azhar et~al.}]{touvron2023llama}
Hugo Touvron, Thibaut Lavril, Gautier Izacard, Xavier Martinet, Marie-Anne Lachaux, Timoth{\'e}e Lacroix, Baptiste Rozi{\`e}re, Naman Goyal, Eric Hambro, Faisal Azhar, et~al. 2023.
\newblock Llama: Open and efficient foundation language models.
\newblock \emph{arXiv preprint arXiv:2302.13971}.

\bibitem[{Vishwakarma et~al.(2024)Vishwakarma, Harkins, Golecha, Bajpe, Dupuis, Buratti, Kremer, Faro, Puri, and Cruz-Benito}]{vishwakarma2024qiskit}
Sanjay Vishwakarma, Francis Harkins, Siddharth Golecha, Vishal~Sharathchandra Bajpe, Nicolas Dupuis, Luca Buratti, David Kremer, Ismael Faro, Ruchir Puri, and Juan Cruz-Benito. 2024.
\newblock {Qiskit} {H}uman{E}val: An evaluation benchmark for quantum code generative models.
\newblock In \emph{2024 IEEE International Conference on Quantum Computing and Engineering (QCE)}, volume~1, pages 1169--1176. IEEE.

\bibitem[{Wang et~al.(2024)Wang, Ma, Zhang, Ni, Chandra, Guo, Ren, Arulraj, He, Jiang et~al.}]{wang2024mmlu}
Yubo Wang, Xueguang Ma, Ge~Zhang, Yuansheng Ni, Abhranil Chandra, Shiguang Guo, Weiming Ren, Aaran Arulraj, Xuan He, Ziyan Jiang, et~al. 2024.
\newblock {MMLU-Pro}: A more robust and challenging multi-task language understanding benchmark.
\newblock \emph{Advances in Neural Information Processing Systems}, 37:95266--95290.

\bibitem[{Yamada et~al.(2025)Yamada, Lange, Lu, Hu, Lu, Foerster, Clune, and Ha}]{yamada2025ai}
Yutaro Yamada, Robert~Tjarko Lange, Cong Lu, Shengran Hu, Chris Lu, Jakob Foerster, Jeff Clune, and David Ha. 2025.
\newblock The ai scientist-v2: Workshop-level automated scientific discovery via agentic tree search.
\newblock \emph{arXiv preprint arXiv:2504.08066}.

\bibitem[{Yang et~al.(2025)Yang, Li, Yang, Zhang, Hui, Zheng, Yu, Gao, Huang, Lv et~al.}]{yang2025qwen3}
An~Yang, Anfeng Li, Baosong Yang, Beichen Zhang, Binyuan Hui, Bo~Zheng, Bowen Yu, Chang Gao, Chengen Huang, Chenxu Lv, et~al. 2025.
\newblock Qwen3 technical report.
\newblock \emph{arXiv preprint arXiv:2505.09388}.

\bibitem[{Ye et~al.(2024)Ye, Wang, Huang, Chen, Zhang, Moniz, Gao, Geyer, Huang, Chen et~al.}]{ye2024justice}
Jiayi Ye, Yanbo Wang, Yue Huang, Dongping Chen, Qihui Zhang, Nuno Moniz, Tian Gao, Werner Geyer, Chao Huang, Pin-Yu Chen, et~al. 2024.
\newblock Justice or prejudice? quantifying biases in llm-as-a-judge.
\newblock \emph{arXiv preprint arXiv:2410.02736}.

\bibitem[{Zheng et~al.(2023)Zheng, Chiang, Sheng, Zhuang, Wu, Zhuang, Lin, Li, Li, Xing et~al.}]{zheng2023judging}
Lianmin Zheng, Wei-Lin Chiang, Ying Sheng, Siyuan Zhuang, Zhanghao Wu, Yonghao Zhuang, Zi~Lin, Zhuohan Li, Dacheng Li, Eric Xing, et~al. 2023.
\newblock Judging {LLM}-as-a-judge with {MT}-{B}ench and {C}hatbot {A}rena.
\newblock \emph{Advances in neural information processing systems}, 36:46595--46623.

\bibitem[{Zhu et~al.(2025)Zhu, Tian, Yang, Zhou, Zhu, Chertkov, Liu, Du, Yuan, Ji et~al.}]{zhu2025probing}
Minhui Zhu, Minyang Tian, Xiaocheng Yang, Tianci Zhou, Penghao Zhu, Eli Chertkov, Shengyan Liu, Yufeng Du, Lifan Yuan, Ziming Ji, et~al. 2025.
\newblock Probing the critical point (critpt) of ai reasoning: a frontier physics research benchmark.
\newblock \emph{arXiv preprint arXiv:2509.26574}.

\bibitem[{Zou et~al.(2025)Zou, Cheng, Aldossary, Bai, Leong, Campos-Gonzalez-Angulo, Choi, Ser, Tom, Wang et~al.}]{zou2025agente}
Yunheng Zou, Austin~H Cheng, Abdulrahman Aldossary, Jiaru Bai, Shi~Xuan Leong, Jorge~Arturo Campos-Gonzalez-Angulo, Changhyeok Choi, Cher~Tian Ser, Gary Tom, Andrew Wang, et~al. 2025.
\newblock El agente: An autonomous agent for quantum chemistry.
\newblock \emph{Matter}, 8(7).

\end{thebibliography}

\end{document}